\documentclass[12pt]{article}

\usepackage{sbc-template}
\usepackage{graphicx,url}
\usepackage[utf8]{inputenc}
\usepackage[table,xcdraw]{xcolor}
\usepackage{multirow}
\usepackage{amsmath}
\usepackage{amssymb}
\title{\textit{verBERT}: Automating Brazilian Case Law Document Multi-label Categorization Using \textit{BERT}}

\author{Felipe R. Serras\inst{1}, Marcelo Finger\inst{1}}

\address{Institute of Mathematics and Statistics (IME) -- University of São Paulo
  (USP)\\
  R. do Matão, 1010 -- Butantã -- São Paulo -- SP -- Brazil -- 05508-090
  \email{\{frserras,mfinger\}@ime.usp.br}
}

\begin{document} 

\maketitle

\begin{abstract}

In this work, we carried out a study about the use of attention-based algorithms to automate the categorization of Brazilian case law documents.  We used data from the \textit{Kollemata} Project to produce two distinct datasets with adequate class systems. Then, we implemented a multi-class and multi-label version of BERT and fine-tuned different BERT models with the produced datasets. We evaluated several metrics, adopting the micro-averaged F1-Score as our main metric for which we obtained a performance value of  $\langle \mathcal{F}_1 \rangle_{micro}=0.72$ corresponding to gains of 30 percent points over the tested statistical baseline.

\end{abstract}

\section{Introduction}
\label{sec:intro}

In this work, we explore the use of \textit{BERT}~\cite{bert} to develop a prototype to automate the categorization of Brazilian case law documents, known as ``\textit{verbetação}''. This categorization has been done manually for the past $25$ years by experts in the field of law, without taking advantage of knowledge engineering techniques; as a result, the categories are not organized as any computationally coherent data structure. Its automation could, therefore, save time and labor, in addition to optimizing its ability to represent knowledge. The problem was brought to us by the \textit{Kollemata} Project team, which organized a library of Brazilian jurisprudence, containing more than 24,000 case law documents from the property law field, and whose dataset was used as the basis for this research.

Unlike other problems in NLP, this one was not clearly associated with a predefined task. We raised several modeling possibilities and  chose multi-label categorization because of the  high value of having an unified set of predetermined thematic categories for this task, manifested both in the literature of the field and by law professionals we consulted.

Considering the scarce literature on this problem, especially in Brazilian Portuguese, our goals were (i) to provide an initial proposal for modeling the problem, compatible with the manifested needs of the consulted professionals, (ii) to assess the potential performance of the main state-of-the-art methods (in this case BERT) on this modeling, and (iii) to map the main obstacles to performance improvement on the adopted modeling. The results of our work are prototypical and seek, mainly, to start the debate and indicate the future paths that research on this topic could take.

To achieve these goals, we reformulated the \textit{Kollemata} dataset with an appropriate class system, and implemented a set of \textit{BERT} classifiers, which can serve as prototypes to automate the categorization of Brazilian case law documents, deepening our understand of the addressed problem and the used algorithm and methodology.

\section{Related Work}
\label{sec:related}

Following the proposition of \textit{BERT}, which included the presentation of the \textit{Multilingual BERT} model, the \textit{BERTimbau} model~\cite{souza2020bertimbau} - the first \textit{BERT} model trained entirely in Brazilian Portuguese - was developed and used in the automation of several tasks in Brazilian Portuguese~\cite{leite2020toxic,souza2019portuguese,salvatore2019logical}.

As far as we know, this is the first work to explore the use of \textit{BERT} to automate case law documents categorization in Brazilian Portuguese. Other works explore the same problem in other languages or dialects of Portuguese~\cite{calambas2015judicial,gonccalves2003preliminary,gonccalves2005linguistic,feinerer2008text,mencia2010efficient}, and one work explores a single-label version of the same problem with less data in Brazilian Portuguese~\cite{de2012clustering}. All of these articles studied algorithms created before \textit{BERT}.

Regarding our efforts to reorganize the data class system, we used clustering techniques to produce an organized class hierarchy, which is not a new ideia for general~\cite{aggarwal2012clustering} or legal applications~\cite{moens2001innovative}. However, here we adopted a methodology designed to meet the specific needs of this task.

Only few recent works relate to ours, both in method and type of task, as the articles of~\cite{villata2020natural} and~\cite{chalkidis2019large}. In the first one, the authors explore different applications of machine learning to Italian law documents, including the use of the \textit{Multilingual BERT} model to perform the categorization of case law documents under a proprietary taxonomy. In the second one, several algorithms, including BERT, were tested to generate a multi-label classifier for European legislative documents in English, based on a large system of topics. These two works are those whose results are most comparable to ours.

\section{Problem}
\label{sec:problem}

\begin{figure}[ht]
\centering
\includegraphics[width=.6\textwidth]{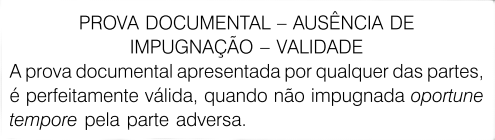}
\caption{A gold standard example of a component of jurisprudence from a Brazilian case law document, extracted from \cite{Augusto2004}.}
\label{fig:CoJ_example}
\end{figure}

In Brazil, case law documents are called ``\textit{acórdãos}" and are used to register the decision of a court and to serve as reference to similar cases in the future (jurisprudence). Searching for these documents is a recurrent task for law professionals and to facilitate it, all case law documents contain a component of jurisprudence (``\textit{ementa jurisprudencial}"), formed by (i) a brief summary of the documents content and (ii) a header, containing a set of descriptive terms that represent the main legal topics covered in the decision. In Portuguese, these terms are called ``\textit{verbetes}" and the header is called ``\textit{verbetação}". Figure \ref{fig:CoJ_example} contains a gold standard example of a Brazilian component of jurisprudence~\cite{Augusto2004}.

The purpose of this component of jurisprudence is to help readers in assessing whether a document is relevant or not to them. However, \cite{Augusto2004} identifies problems in the ways these headers are produced, including redundancy, usage of generic terms, lack of a consistent vocabulary, and lack of well-defined syntactic structures, both within the terms and between them. These problems affect the appropriate usage of the component of jurisprudence and get in the way of having a well formed and nationally validated set of case law descriptor terms, seen as an important asset within the Brazilian theory of law.

As stated in Section \ref{sec:intro}, given the importance of having this set of predetermined descriptor terms, we decided to model it as a task of multi-class and multi-label text categorization i.e. the task of assigning to each document $d_j$ in a set of natural language documents $\mathcal{D}$, a variable number of categories $c_i$, taken from a set of multiple possible predefined categories $\mathcal{C}$~\cite{Sebastiani2002}. Also, fitting the data to this model could serve as basis for the resolution of several problems identified by~\cite{Augusto2004}.

\section{Materials and Methodology}
\label{sec:matmet}
\subsection{\textit{Kollemata} and \textit{verBERT}}

We used data from the \textit{Kollemata} Project\footnote{\url{https://www.kollemata.com.br/}}. Its database contains more than 24,000 entries, each one corresponding to one case law document from the property law field and containing its summary and header\footnote{The \textit{Kollemata} Project dataset was given to our research by the \textit{Kollemata} Project team. The data set will not be publicly available for now, as Brazilian authorities are reviewing the correct way to make these documents available on a large scale without exposing sensitive data.}.

Our goal was to use the pairs of summary and header content as the examples to BERT fine-tuning, aiming to automatically generate the terms of the header from the summary content. The procedures performed to prepare this data for this usage are explained in Section \ref{sec:dataset}.

We implemented a \textit{Python3} program called \textit{verBERT}, that uses the Hugging Face Transformers Package \cite{wolf2019huggingface} to fine-tune, evaluate and test \textit{BERT} multi-label classifiers using pre-treined \textit{BERT} models. In doing so, we refactored, adapted and incremented the examples of Hugging Face for generating BERT Classifiers over GLUE~\cite{wang2019glue}, with copyright from Hugging Face, Google and Nvidia, but available under the Apache 2.0 License\footnote{The \textit{verBERT} source code, as well as other resources, are available in our repository: \url{https://github.com/frserras/verbert-categorization}}. 

Our multi-label categorization architecture was inspired by~\cite{multibert} and uses a simple feed-forward network coupled with sigmoid functions $\sigma$ as the categorization layer (Equations \ref{eq:class_layer} and \ref{eq:sigma}, where $\mathbf{a}$ is the \textit{BERT} output representation, $W^{C}: W^{C} \in \mathbb{R}^{d \times |\mathcal{C}|}$ a linear transformation matrix, $\mathbf{b}$ a bias vector, and $d$ is the characteristic dimension of the model)~\cite{han1995influence}, and the mean of the binary cross-entropy ~\cite[p. 65-66]{Goodfellow-et-al-2016} over the labels as the loss function to be minimized using Adam optimizer ~\cite[p. 305-306]{Kalchbrenner2014,Goodfellow-et-al-2016}  with weight decay regularization~\cite{loshchilov2017decoupled}.

\begin{equation}
    \label{eq:class_layer}
    \mathbf{c}=\varsigma(\mathbf{a}W^{C}+\mathbf{b})
\end{equation}
\begin{equation}
    \label{eq:sigma}
    \varsigma(\mathbf{x}=\{x_1,...,x_n\}) = \{\sigma(x_1),...,\sigma(x_n)\}
\end{equation}

We evaluated and compared three pre-trained \textit{BERT} models as bases for our classifier generation: \textit{Multilingual BERT}, a base model trained over $104$ languages on Wikipedia data~\cite{bert}, and the two variants (base and large) of the \textit{BERTimbau} Model, pre-trained entirely in Portuguese~\cite{souza2020bertimbau,souza2019portuguese}.

% using data from the \textit{brwac dataset}~\cite{wagner2018brwac}.

\subsection{Metrics and Baseline}
To compare the models produced by \textit{verBERT}, we adopted three performance metrics commonly used in problems of categorization and information retrieval: precision $\mathcal{P}$, recall $\mathcal{R}$, and F1-Score $\mathcal{F}_1$~\cite{godbole2004discriminative}. For each metric $\mathcal{E}$ we computed its three averaged multi-label versions: the micro-average $\langle \mathcal{E} \rangle_{micro}$, the macro-average $\langle \mathcal{E} \rangle_{macro}$ and the average by instance $\langle \mathcal{E} \rangle_{inst}$ ~\cite{Koyejo2015}

We also used the accuracy metric~\cite{godbole2004discriminative}, for which we selected two versions for the multi-label case: the Hamming accuracy $\mathcal{A}$ i.e. the percentage of labels correctly predicted over all examples, and the subset accuracy $\dot{\mathcal{A}}$, the percentage of examples for which all labels were correctly predicted.

We selected the $\langle \mathcal{F}_1 \rangle_{micro}$ as our main metric, since it tends to be more appropriate for information retrieval cases with strong unbalanced categories, like ours. As secondary metrics, we adopted (i) the $\langle \mathcal{F}_1 \rangle_{macro}$, which shows us the impact of category imbalance on performance, if compared with its micro-averaged counterpart, and (ii) the subset accuracy $\dot{\mathcal{A}}$, because is a more robust metric for problems of categorization, that serves as a lower bound to our performance.

Besides the models produced by \textit{verBERT}, we implemented a simple statistical baseline method, where the $n=5$ most common categories of the dataset were used to categorize all of its entries\footnote{We tested different values for $n$ and choose the one that maximized the baseline performance for our main metrics.}. Its results are presented in Table \ref{tab:comp_final}. This type of baseline, despite its simplicity, is a robust option for unbalanced category systems, where betting on the most likely categories can be a good strategy for blindly maximizing performance.

\section{Data preparation and Ontological Adjustment}
\label{sec:dataset}

We used the pairs [summary, header], from the \textit{Kollemata} Project dataset and pre-processed them by (i) filtering corrupted \textit{HTML} notation from the summaries, (ii) unifying the terms for law entities across the dataset, and (iii) mapping and unifying different segmentation symbols between descriptor terms. The data was structured in a list of entries, each one corresponding to a case law document and containing the summary and an ordered list of its header descriptor terms.

We then performed an exploratory analysis on the dataset, to identify its main properties and eventual obstacles for automation. We evaluated the distributions, across the data, of summary size,  number of categories per header, presence of the descriptor terms in their respective summary, and category occurrence.

From the summary size distribution, we observed that the expected number of sub-words by summary is less than the internal BERT dimensionality $d$ for our three models, which confirms that this configuration is in the range where BERT is expected to be more efficient than recurrent models~\cite{bert}.

The categories per header distribution revealed an average of $5$ descriptor terms per document, while the distribution of presence of the descriptor terms in their respective summary revealed that most summaries contain a fraction of their respective descriptor terms, but not all. Both of these results characterize a scenario compatible with modeling the problem as a multi-label classification task.

On the other hand, the category occurrence distribution revealed a total number of categories (20,780) close to the number of documents in the dataset, most of them with low occurrence. This is due to the high specificity of the descriptor terms, that contain many concepts and sub-concepts within themselves. The descriptor terms ``\textit{ineficácia da adjudiação}", ``\textit{ineficácia da alienação}" and ``\textit{ineficácia da caução}" are a good example. All of them represent, at the same time, the concept of ``\textit{ineficácia}" and another sub-concept, unique to each of these terms. Ideally, the concept and sub-concept would be separated somehow.

This corroborates the problems pointed out by \cite{Augusto2004}, highlighted in Section \ref{sec:problem}. To enable the treatment of the problem via multi-label categorization, we performed an ontological adjustment step, in an attempt to extract the concepts contained in each descriptor term and organize them in a class system with reasonable size.

Before that, however, we also examined the correlation between the orders of appearance of any two terms in the headers, expecting to detect any type of underlying hierarchy between them that could be used in our ontological adjustment. We observed a strong pairwise correlation, but not to the point of building longer and more complex hierarchical lineages between the categories. From this we concluded that segmenting the categories in their components and then build a hierarchical structure from them, taking advantage of these co-occurrences was the best way to adequate the class system.
 
During this ontological adjustment, category descriptor terms were tokenized in its component words, stop words were removed using the NLTK\footnote{\url{https://www.nltk.org/}} Portuguese stop words list, and then stemmed with a manually enriched version of the RSLP Stemmer~\cite{huyck2001stemming}. We removed the terms with less then 5 occurrences to eliminate spurious terms and then assigned paternity relations between terms with high co-occurrence.

The terms in the top of the obtained hierarchy with low occurrence were grouped under the ``Others'' category, while the ones with high occurrence were automatically clustered in $25$ super-categories, each connected to the component terms of the original categories by the hierarchical structure.

We clustered the categories using a K-means clusterizer applied to arrays with identifiers of the documents in which each category appeared\footnote{This clustering was performed as a pre-processing step. The classifier performs the categorization on already clustered data and therefore has no access to the arrays with identifiers of the documents in which each category appeared, which could introduce a bias towards the actual labels.}~\cite[p. 515-518]{manning1999foundations}. These arrays had their dimension previously reduced to $d=50$ using the \textit{scikit-learn} Trucated SVD Algorithm\footnote{\url{https://scikit-learn.org/stable/modules/generated/sklearn.decomposition.TruncatedSVD.html}}. We tested different values of reduced dimension $d$ and different clustering methods. Each combination resulted in different amounts of clusters, and the chosen combination was the one whose clustering groups were the most cohesive.

Despite being justified by the properties of the data, the ontological adjustment carried out here is extensive. For this reason, the performance values obtained should be understood as the joint performance between the categorization method and the adopted data reorganization methodology.

\section{Experiments and Results}
\label{sec:exps}

To evaluate and compare the performance of the models produced by \textit{verBERT} and our hierarchy of categories, we executed a series of experiments. We used  three pre-trained models, \textit{Multilingual BERT}, \textit{BERTimbau} base, and \textit{BERTimbau} large; and two different versions of the produced hierarchy and dataset: dataset 1, produced with a smaller grouping rate and maintaining the ``Others'' category, and dataset 2, a slightly more homogeneous option, produced with higher grouping rate and excluding the ``Others'' category.

Each combination of model and dataset prompted a group of experiments, in which we varied the maximum learning rate $\lceil \eta \rceil$, reached after $50$ linear warming-up steps ($2 \cdot 10^{-5}$, $4 \cdot 10^{-5}$, $5 \cdot 10^{-5}$, $1 \cdot 10^{-4}$, $1 \cdot 10^{-3}$, $5 \cdot 10^{-4}$, $5 \cdot 10^{-3}$, and $1 \cdot 10^{-2} $)\footnote{We used learning rate values based on the ones used in~\cite{bert} and~\cite{souza2019portuguese}, and included larger values to get a more general picture of the performance evolution.}, the  maximum input sentence size $|\mathcal{S}|$ ($52$, $68$, $131$, and $200$ tokens) and the categorization threshold probability $P_{ct}$, above which a category is assigned to a document by the model ($0.25$, $0.50$ and $0.75$).

We divided each dataset in a training set ($72\%$), a validation set ($8\%$) and a test set ($20\%$). In each experiment the model was fine-tuned with the training set for $10$ epochs and evaluated at regular intervals against the validation set. The intermediate model with the best performance on the validation set was then evaluated on the test set.

For that, we used two machines: (i) one with $48$ cores, $378$ GB of primary memory and a \textit{NVIDIA Tesla K40c GPU} ($11.4$ GB), and the other (ii) with $40$ cores, $62.8$ GB of primary memory, and a \textit{NVIDIA GeForce GTX1080Ti GPU} ($11.7$ GB). Since both machines had secondary memory limitations, \textit{verBERT} was planned to use little storage as possible.

Comparing the results of the experiments, we observe an improvement in performance with higher values of maximum learning rate $\lceil \eta \rceil$ and input sentence size $|\mathcal{S}|$.The increase in performance with input sentence size is observed for all values of $|\mathcal{S}|$, but the increase rate decreases progressively, indicating saturation of this trend. On the other hand, the growth in performance with maximum learning rate $\lceil \eta \rceil$ is observed only up to $\lceil \eta \rceil=5 \cdot 10^{-4}$, from which point on, performance rates fall sharply. Our investigations suggest that this is due to a local minimum, where the optimizer got stuck.

We also observed that the categorization threshold probability dictates the optimization strategy adopted by the algorithm: with higher values of $P_{ct}$, it is easier to maximize precision first and then slightly decrease it to increase recall and F1-score. In contrast, with smaller threshold probability values, it is easier to maximize recall first and then decrease it optimally, to enhance precision and F1-score.

To provide a global view, Table \ref{tab:dataset_model_compared} presents the main performance metrics in the best case for each group of experiments, and the parameters for which these results were obtained.

% \usepackage{multirow}
\begin{table}[]
\centering
\footnotesize
\caption[]{Best results for the main metrics and respective parameters from each group of experiments on the test set.}
\label{tab:dataset_model_compared}
\begin{tabular}{|c|c|c|}
\hline
\rowcolor[HTML]{EFEFEF} 
\multicolumn{1}{|l|}{\cellcolor[HTML]{EFEFEF}$\downarrow$Dataset/Model$\rightarrow$} & \multicolumn{2}{c|}{\cellcolor[HTML]{EFEFEF}\textbf{BERTimbau LARGE}}                                                       \\ \hline
                                                                                    & $\langle \mathcal{F}_1 \rangle_{micro} =0.71$ & $\dot{\mathcal{A}} =0.24$                                                 \\ \cline{2-3} 
\multirow{-2}{*}{Dataset 1}                                                         & $\langle \mathcal{F}_1 \rangle_{macro} =0.66$ & $(P_{limiar}=0.25;\lceil \eta \rceil=1 \cdot 10^{-4}; |\mathcal{S}|=200)$   \\ \hline
                                                                                    & $\langle\mathcal{F}_1 \rangle_{micro} =0.72$  & $\dot{\mathcal{A}} =0.38$                                                 \\ \cline{2-3} 
\multirow{-2}{*}{Dataset 2}                                                         & $\langle\mathcal{F}_1 \rangle_{macro} =0.71$  & $(P_{limiar}=0.50;\lceil \eta \rceil =1 \cdot 10^{-4}; |\mathcal{S}| =131)$ \\ \hline
\rowcolor[HTML]{EFEFEF} 
\multicolumn{1}{|l|}{\cellcolor[HTML]{EFEFEF}$\downarrow$Dataset/Model$\rightarrow$} & \multicolumn{2}{c|}{\cellcolor[HTML]{EFEFEF}\textbf{Multilingual BASE}}                                                     \\ \hline
                                                                                    & $\langle\mathcal{F}_1 \rangle_{micro} =0.69$  & $\dot{\mathcal{A}} =0.24$                                                 \\ \cline{2-3} 
\multirow{-2}{*}{Dataset 1}                                                         & $\langle\mathcal{F}_1 \rangle_{macro} =0.59$  & $(P_{limiar}=0.50;\lceil \eta \rceil=1 \cdot 10^{-4}; |\mathcal{S}|=200)$   \\ \hline
                                                                                    & $\langle\mathcal{F}_1 \rangle_{micro} =0.71$  & $\dot{\mathcal{A}} =0.37$                                                 \\ \cline{2-3} 
\multirow{-2}{*}{Dataset 2}                                                         & $\langle\mathcal{F}_1 \rangle_{macro} =0.70$  & $(P_{limiar}=0.50;\lceil \eta \rceil=5 \cdot 10^{-5}; |\mathcal{S}|=200)$   \\ \hline
\rowcolor[HTML]{EFEFEF} 
\multicolumn{1}{|l|}{\cellcolor[HTML]{EFEFEF}$\downarrow$Dataset/Model$\rightarrow$} & \multicolumn{2}{c|}{\cellcolor[HTML]{EFEFEF}\textbf{BERTimbau BASE}}                                                        \\ \hline
                                                                                    & $\langle\mathcal{F}_1 \rangle_{micro} =0.70$  & $\dot{\mathcal{A}} =0.24$                                                 \\ \cline{2-3} 
\multirow{-2}{*}{Dataset 1}                                                         & $\langle\mathcal{F}_1 \rangle_{macro} =0.59$  & $(P_{limiar}=0.50;\lceil \eta \rceil=1 \cdot 10^{-4}; |\mathcal{S}|=200)$   \\ \hline
                                                                                    & $\langle\mathcal{F}_1 \rangle_{micro} =0.72$  & $\dot{\mathcal{A}} =0.38$                                                 \\ \cline{2-3} 
\multirow{-2}{*}{Dataset 2}                                                         & $\langle\mathcal{F}_1 \rangle_{macro} =0.70$  & $(P_{limiar}=0.25;\lceil \eta \rceil=4 \cdot 10^{-5}; |\mathcal{S}|=200)$   \\ \hline
\end{tabular}
\end{table}

We observed better results on average for the large model, than for the two base models, for both datasets. However, this improvement is low: only two percent point increase in $\langle \mathcal{F}_1 \rangle_{micro}$ for the best case. This seems to indicate the superiority of the larger model and the robustness of the smaller ones at the same time. This improvement is larger for macro-F1 over the dataset 1. Since the dataset 1 is less homogeneous and the macro-F1 is a metric more sensitive to the lack of homogeneity, this suggests that the large model is better at dealing with less homogeneous category systems. 

Comparing the results for the \textit{BERTimbau} models with the results for the \textit{Multilingual} model, we see that using Portuguese based models improves average performance, but again, this improvement is low, what may indicate the robustness of the \textit{Multilingual} model.

Looking at the impact of data homogeneity over performance, we see that the use of dataset 2 instead of dataset 1 brings an improvement to all models, specially in the subset accuracy $\dot{\mathcal{A}}$, which indicates a significant increase in our performance lower bound. Furthermore, we also see an increase in the macro-averaged F1, which reveals that a small improvement in homogeneity was enough to cause a great impact in the gap between the micro-averaged and the macro-averaged F1 scores. These results should nonetheless be interpreted with caution, given the difference between the test sets of dataset 1 and dataset 2 . Also, with dataset 2 we obtained our best model, the $verBERT \star$, obtained by fine-tuning \textit{BERTimbau} large with $P_{ct} = 0.5$, $\lceil \eta \rceil= 1 \cdot 10^{-4}$ and $|\mathcal{S}|=131$. In Table \ref{tab:comp_final} we compare all the performance metrics obtained by $verBERT \star$ and by our baseline, described in Section \ref{sec:matmet}

\begin{table}[]
\centering
\footnotesize
\caption[]{A comparison of the performance results obtained  by the $verBERT\star$ model and by our statistical baseline, for all performance metrics.}
\label{tab:comp_final}
\begin{tabular}{l|l|l|l|l|l|ll}
\cline{2-7}
                                                              & \cellcolor[HTML]{EFEFEF}$\langle \mathcal{P} \rangle_{micro}$   & \cellcolor[HTML]{EFEFEF}$\langle \mathcal{P} \rangle_{macro}$   & \cellcolor[HTML]{EFEFEF}$\langle \mathcal{P} \rangle_{inst}$   & \cellcolor[HTML]{EFEFEF}$\langle \mathcal{R} \rangle_{micro}$ & \cellcolor[HTML]{EFEFEF}$\langle \mathcal{R} \rangle_{macro}$ & \multicolumn{1}{l|}{\cellcolor[HTML]{EFEFEF}$\langle \mathcal{R} \rangle_{inst}$} &  \\ \cline{1-7}
\multicolumn{1}{|l|}{\cellcolor[HTML]{EFEFEF}Baseline}        & 0.33                                                            & 0.06                                                            & 0.33                                                           & 0.47                                                          & 0.19                                                          & \multicolumn{1}{l|}{0.49}                                                         &  \\ \cline{1-7}
\multicolumn{1}{|l|}{\cellcolor[HTML]{EFEFEF}$verBERT \star$} & 0.77                                                            & 0.77                                                            & 0.79                                                           & 0.67                                                          & 0.66                                                          & \multicolumn{1}{l|}{0.73}                                                         &  \\ \cline{1-7}
                                                              & \cellcolor[HTML]{EFEFEF}$\langle \mathcal{F}_1 \rangle_{micro}$ & \cellcolor[HTML]{EFEFEF}$\langle \mathcal{F}_1 \rangle_{macro}$ & \cellcolor[HTML]{EFEFEF}$\langle \mathcal{F}_1 \rangle_{inst}$ & \cellcolor[HTML]{EFEFEF}$\mathcal{A}$                         & \cellcolor[HTML]{EFEFEF}$\dot{\mathcal{A}}$                 &                                                                                   &  \\ \cline{1-6}
\multicolumn{1}{|l|}{\cellcolor[HTML]{EFEFEF}Baseline}        & 0.39                                                            & 0.09                                                            & 0.37                                                           & 0.80                                                          & 0.00                                                          &                                                                                   &  \\ \cline{1-6}
\multicolumn{1}{|l|}{\cellcolor[HTML]{EFEFEF}$verBERT \star$} & 0.72                                                            & 0.71                                                            & 0.73                                                           & 0.95                                                          & 0.38                                                          &                                                                                   &  \\ \cline{1-6}
\end{tabular}
\end{table}

\section{Conclusions}
\label{sec:concl}


The $verBERT \star$ model achieves results significantly superior to the baseline for all metrics. Beyond that, the value obtained by $verBERT \star$ for our main metric, the micro-averaged F1 score, is close to the $0.732$ value obtained by~\cite{chalkidis2019large}, and is higher then the $0.505$ value, obtained by~\cite{villata2020natural} \footnote{As mentioned in Section \ref{sec:related}, these two works are those whose methods, applications, and therefore results, are most comparable to ours.}.

These results are quite positive given the difficulties in modeling the problem, and indicate that our approach is promising. However, (i) a direct comparison would be inappropriate, given the methodological differences between our work and those cited, (ii) the observed performance is still far from the optimal values obtained for other applications in NLP, and (iii) to obtain such a level of performance it was necessary to carry out a heavy pre-processing in which we observed several of the taxonomic problems of the generation of headers, pointed out in the literature. This indicates that the lack of standardization in headers is the main obstacle to achieving better performance results.

Despite the usefulness of our models as categorization prototypes that could already be used to assist human classifiers, we believe that, given these results, better performance values could only be obtained (i) by exploring models other than the categorization, which would compromise the taxonomic value of the descriptor terms pointed by field specialists, or (ii) by revisiting the production system of the headers in its core, in order to manually generate a system of categories that is appropriate for the main activity and that could be used in the automation process without so much pre-processing . In addition, (iii) repetitions of our experiments could bring greater statistical value to these results, mitigating the uncertainty brought by the differences between the class systems of our two datasets, and (iv) the exploration of more robust baselines based on other natural language processing techniques could help us better understand the quality of the performance obtained by our method. These are the main issues we would like to address in future works. 

\section*{Acknowledgements}

This work was carried out  as a master's research project \cite{Serras2021Algoritmos} at the Center for Artificial Intelligence (C4AI-USP), with
support by the São Paulo Research Foundation (FAPESP grant \#2019/07665-4) and by the IBM Corporation. This study was financed in part by the Coordenação de Aperfeiçoamento de Pessoal de Nível Superior -- Brasil (CAPES) -- Finance Code 001. We would like to thank Dr. Sérgio Jacomino, his team from the \textit{Kollemata Project}, Professors Adriana Unger,  Juliano Maranhão, Renata Vieira, Denis Mauá and the anonymous reviwers for their help.

\bibliographystyle{sbc}
\bibliography{bibliography}

\end{document}